\documentclass[10pt,letterpaper]{article}
\usepackage[margin=1in]{geometry}
\usepackage{natbib}
\usepackage{amsmath,amssymb,amsthm}
\usepackage{graphicx}
\usepackage{booktabs}
\usepackage{multirow}
\usepackage{xcolor}
\usepackage{subcaption}
\usepackage{url}
\usepackage{tikz}
\usetikzlibrary{arrows.meta,positioning,calc,fit,backgrounds}
\graphicspath{{figs/}}

\title{Cross-Trajectory Chimera Interventions Reveal Dissociable Roles of Weight Magnitude and Direction in Grokking}

\author{Truong Xuan Khanh\\H\&K Research Studio, Clevix LLC, Hanoi, Vietnam\\\texttt{khanh@clevix.vn}}

\begin{document}
\maketitle

\begin{abstract}
Which properties of a partially trained network are causally portable to a
\emph{different}, independently trained network, and which are not?
Interventions on a single training trajectory cannot answer this: they can
show that a property is \emph{necessary} for an outcome within one run, but not
whether it is \emph{portable} across runs. We introduce \emph{cross-trajectory
chimera interventions}: given two networks trained from different random seeds
on the same task, we decompose each weight vector into a magnitude (norm) and a
direction (unit vector), recombine the magnitude of one run with the direction
of another to form a \emph{chimera}, and continue training. Across two
modular-arithmetic tasks that exhibit grokking, we find a clean dissociation. The \emph{direction} carries a
transferable component associated with the final circuit's spectral identity:
implanting a donor's direction while keeping the recipient's norm drives the
continued run toward the donor's eventual circuit in $40/40$ independent
recombinations, and this transfer is specific to the donor's content rather than
to the mere angular displacement (an angle-matched random control produces no
such shift). Interpolating the direction along the geodesic between two runs
reveals that this transfer is \emph{threshold-like} rather than a continuous
blend. Most notably, cross-trajectory intervention reveals that transferability
itself depends on the recipient's dynamical state: the interpolation threshold
at which identity flips from recipient-like to donor-like is predicted by the
recipient's weight norm, separating perfectly by norm class across all $20$
pairs and both tasks (joint exact permutation probability
$1.9{\times}10^{-4}$). We localize the threshold to a resolution of
$\pm 1/64$ using an adaptive bisection procedure that we contribute as a
reusable measurement tool for stability-under-protocol interventions. By
contrast, the \emph{norm} carries only a modest, spatially distributed effect on
grokking delay and no measurable identity information. Together these results
map a concrete geometric division of labour: direction indexes which solution a
trajectory approaches, while norm governs how susceptible that identity is to
being overwritten.
\end{abstract}

\section{Introduction}
\label{sec:intro}

Grokking---the delayed onset of generalization long after a network has fit its
training data---has become a controlled setting for asking how neural networks
transition from memorization to structured solutions
\citep{power2022grokking,nanda2023progress}. A large body of recent work
intervenes on a single training run to probe this transition: freezing,
rescaling, or projecting weights along a trajectory and observing the effect on
delay or on the emerging circuit
\citep{nanda2023progress,varma2023explaining}. Such within-trajectory
interventions, however, cannot answer a distinct question: of everything a
partially trained checkpoint contains, which parts are \emph{portable}---able to
determine an outcome when transplanted into a \emph{different}, independently
trained run? Within-trajectory interventions can establish that a property is
\emph{necessary} for an outcome inside one run; they cannot establish whether
that property is \emph{portable}---able to produce the outcome when placed in a
different run's context. Distinguishing these two questions is our organizing
concern throughout the paper: we do not ask why direction encodes circuit
identity, but whether it is causally portable, and if so, under what
conditions.

We study this question through a decomposition of the weight vector into a
radial part (its norm) and an angular part (its unit direction),
$\theta = r \cdot u$. This decomposition itself is not new: the interplay of
weight norm and direction in grokking has been analyzed within single
trajectories \citep{liu2022towards,varma2023explaining}. Our contribution is a
\emph{cross-trajectory} causal probe. Given two networks $A$ and $B$ trained
from different seeds, we build a \emph{chimera} by combining $B$'s norm with
$A$'s direction (or vice versa), continue training, and ask two questions: does
the grokking \emph{delay} of the chimera follow the run that donated the norm,
and does the final \emph{circuit identity} follow the run that donated the
direction?

We find a sharp asymmetry. Circuit identity---measured by a task-appropriate
power spectrum of the token embedding---follows the angular donor cleanly,
consistently, and specifically to that donor's content. Grokking delay follows
the radial donor only weakly, and this weak effect cannot be localized to any
single layer group. Pushing further, we interpolate the implanted direction
along the geodesic from $A$ to $B$ at fixed norm and find that identity transfer
is \emph{threshold-like}: the final circuit remains recipient-like across a
plateau of interpolation values and then switches to donor-like over a narrow
range. The location of this threshold is not constant---and, revealingly, it is
predicted by the recipient's weight norm. This last finding is only accessible
\emph{because} the intervention is cross-trajectory: it concerns not merely what
transfers, but when transfer is possible.

\paragraph{Contributions.}
\begin{enumerate}
\item \textbf{Method.} Cross-trajectory chimera interventions, a causal probe
that recombines the magnitude and direction of independently trained networks,
together with an angle-matched random control that isolates donor-specific
content from generic angular perturbation (Section~\ref{sec:methods}).
\item \textbf{Dissociation.} The angular component carries a transferable,
donor-specific component associated with circuit identity across two tasks
($40/40$ recombinations sign-correct); the radial component carries only a
modest, non-localizable delay effect (Section~\ref{sec:endpoints},
\ref{sec:delay}).
\item \textbf{Threshold and its dependence on state.} Identity transfer is
threshold-like in the interpolation coordinate, and the threshold location is
predicted by the recipient's norm, separating perfectly by norm class across all
$20$ pairs and both tasks. We contribute an adaptive bisection procedure that
localizes the threshold to $\pm 1/64$ at roughly three evaluations per pair, a
reusable measurement strategy for interventions that are stable under their
training protocol (Section~\ref{sec:threshold}); this separation, and the
identity-transfer result above, both survive an optimizer-state ablation that
transplants Adam moments between donor and recipient rather than resetting them
(Section~\ref{sec:optablation}).
\item \textbf{Supporting characterization.} Circuit spectra continue
reorganizing for hundreds to thousands of steps after generalization is
behaviourally complete, and no two seeds converge to near-identical spectra
(Section~\ref{sec:supporting}).
\end{enumerate}

\section{Related Work}
\label{sec:related}

\paragraph{Transferring weights between grokking runs.}
Closest to our setting is work that transfers a learned embedding from a weaker,
already-generalized network to initialize a stronger one, accelerating or
inducing grokking in the target \citep{groktransfer2025}. That line treats
transfer as a method for \emph{speeding up} generalization and measures its
effect on delay and accuracy. Our aims differ in three ways. First, we recombine
two \emph{peer} runs by explicitly separating norm from direction, rather than
transplanting a whole embedding from a privileged source. Second, our primary
outcome is \emph{circuit identity}, not delay; nothing in the acceleration line
measures which of several distinct circuits the recipient adopts. Third, we
introduce a donor-content control (an angle-matched random direction) that
distinguishes ``$B$'s specific direction mattered'' from ``a perturbation of
this angular size mattered.'' We also find that the identity-carrying signal is
present in multiple hidden layer groups, not the embedding alone
(Section~\ref{sec:supporting}), which situates our result outside an
embedding-centric account.

\paragraph{Merging independently trained networks.}
A parallel literature merges or interpolates independently trained networks
after aligning their permutation symmetries, seeking low-loss connectivity or a
single merged model \citep{ainsworth2022git,entezari2021role}. We deliberately
do \emph{not} align permutations: the point of a chimera is to serve as a causal
probe, and we measure a circuit-identity signature and the dynamics of continued
training rather than a loss barrier between fixed endpoints.

\paragraph{Basin selection and mode determinism.}
Which solution a trajectory reaches has been studied through linear mode
connectivity and the role of the early training phase in determining the
eventual basin \citep{frankle2020linear,fort2020deep}, and more recently through
singular-learning-theory and exploration-based accounts of basin selection in
grokking \citep{basin2026slt}. These analyses are largely
observational---they characterize which basin is reached under natural training.
Our chimera intervention provides a causal complement: by transplanting a
direction across trajectories we measure the boundary at which basin membership
changes, and we show that the location of that boundary depends on the
recipient's state.

\paragraph{Relationship to our prior work.}
Two of our earlier studies analyze single-trajectory grokking: one establishes
that norm and direction are jointly necessary to explain the delay within a run,
and the other calibrates a norm--timescale relationship for the delay. The
present paper is neither an extension nor a re-analysis of those datasets. It
concerns \emph{cross-trajectory} transferability and a dissociation between the
two components' portable content, using disjoint experiments; where we note that
the modest delay effect is consistent with the earlier norm--timescale
relationship, we say only that, and draw no stronger link.

\section{Setup and Methods}
\label{sec:methods}

\paragraph{Tasks and model.}
We study two tasks on which a small transformer groks: modular addition
$(a+b)\bmod p$ and modular multiplication $(a\cdot b)\bmod p$ on the
multiplicative group $\{1,\dots,p-1\}$, with $p=59$. Inputs are tokenized as
$[a,b,{=}]$ and the model predicts the result token. The architecture is a
one-layer transformer (embedding dimension $128$) trained full-batch with AdamW
($\text{lr}=10^{-3}$, weight decay $1.0$, $\beta=(0.9,0.98)$) at a $50\%$
train/test split. For multiplication we restrict to the multiplicative group
because including $0$ makes every $(0,b)$ pair collapse to $0$, a degenerate
shortcut that both inflates accuracy and pollutes the spectral fingerprint
defined below.

\paragraph{Radial--angular decomposition and chimeras.}
For a flattened weight vector $\theta$ we write $\theta = r\,u$ with $r =
\lVert\theta\rVert$ and $u = \theta/\lVert\theta\rVert$. Given a pre-grokking
checkpoint from run $A$ (recipient) and one from run $B$ (donor), we form
\emph{chimera} variants and continue training under the recipient's optimizer
state; results are reported for a reset optimizer unless noted, and we verify in
Section~\ref{sec:optablation} that this choice does not drive our conclusions.
The core
variant is \textsc{radial} ($r_B\,u_A$: recipient direction, donor norm) and its
mirror \textsc{reverse\_radial} ($r_A\,u_B$). We include four controls:
\textsc{mid\_norm} ($\tfrac{1}{2}(r_A{+}r_B)\,u_A$, isolating the effect of
changing norm alone), \textsc{random\_u} (a fully random direction at the
recipient's norm), and---critically---\textsc{matched\_random} (a random
direction placed at \emph{exactly} the angular distance that $u_B$ has from
$u_A$). In high dimensions a uniformly random direction is nearly orthogonal to
$u_A$, whereas two independently trained runs on the same task are not; the
matched-random control equalizes the angular displacement so that any difference
from the true donor isolates donor-specific content rather than perturbation
magnitude.

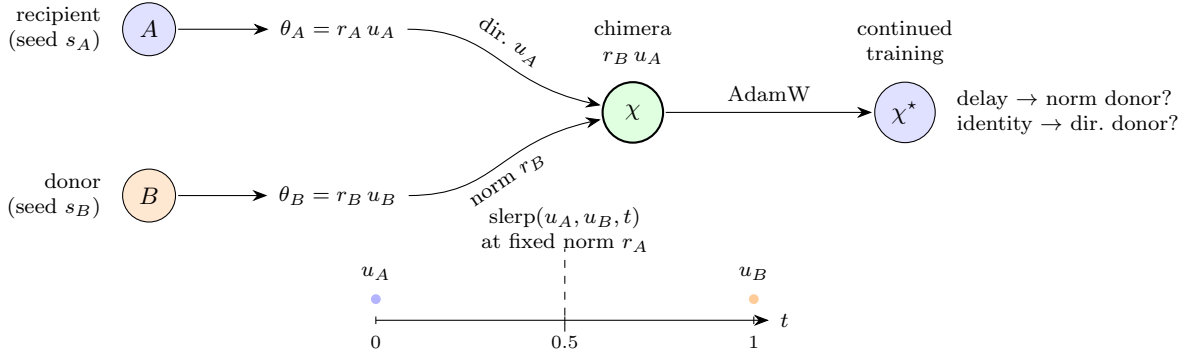
\begin{figure}[t]
\centering
\begin{tikzpicture}[
    >={Stealth[length=2mm]},
    run/.style={circle, draw, minimum size=7mm, inner sep=0pt, font=\small},
    chim/.style={circle, draw, thick, minimum size=8mm, inner sep=0pt, font=\small},
    lbl/.style={font=\footnotesize},
    every node/.style={outer sep=1pt},
]
\node[run, fill=blue!12] (A) at (0,1.1) {$A$};
\node[run, fill=orange!18] (B) at (0,-1.1) {$B$};
\node[lbl, left=1mm of A, align=right] {recipient\\(seed $s_A$)};
\node[lbl, left=1mm of B, align=right] {donor\\(seed $s_B$)};
\node[lbl] (dA) at (2.5,1.1) {$\theta_A = r_A\, u_A$};
\node[lbl] (dB) at (2.5,-1.1) {$\theta_B = r_B\, u_B$};
\draw[->] (A) -- (dA);
\draw[->] (B) -- (dB);
\node[chim, fill=green!12] (C) at (6.4,0) {$\chi$};
\node[lbl, above=0.5mm of C, align=center] {chimera\\$r_B\, u_A$};
\draw[->] (dA) .. controls (4.6,1.1) and (4.6,0.4) .. (C)
    node[lbl, midway, above, sloped] {dir.\ $u_A$};
\draw[->] (dB) .. controls (4.6,-1.1) and (4.6,-0.4) .. (C)
    node[lbl, midway, below, sloped] {norm $r_B$};
\node[run, fill=blue!12, minimum size=8mm] (F) at (10.0,0) {$\chi^\star$};
\node[lbl, above=0.5mm of F, align=center] {continued\\training};
\draw[->] (C) -- (F) node[lbl, midway, above] {AdamW};
\node[lbl, right=1mm of F, align=left] {delay $\to$ norm donor?\\identity $\to$ dir.\ donor?};
\begin{scope}[shift={(3.0,-2.75)}]
  \draw[->] (0,0) -- (5.2,0) node[right, lbl] {$t$};
  \foreach \x/\t in {0/0, 2.5/0.5, 5/1} \draw (\x,0.05)--(\x,-0.05) node[below, font=\scriptsize]{\t};
  \node[circle,fill=blue!30,inner sep=1.3pt] at (0,0.28){};
  \node[lbl] at (0,0.62) {$u_A$};
  \node[circle,fill=orange!40,inner sep=1.3pt] at (5,0.28){};
  \node[lbl] at (5,0.62) {$u_B$};
  \draw[dashed] (2.5,-0.15) -- (2.5,0.95);
  \node[lbl, align=center] at (2.5,1.2) {$\mathrm{slerp}(u_A,u_B,t)$\\at fixed norm $r_A$};
\end{scope}
\end{tikzpicture}
\caption{\textbf{Cross-trajectory chimera intervention.} Two networks are trained
from different seeds on the same task. Each weight vector is split into a norm
$r$ and a unit direction $u$. A \emph{chimera} recombines one run's norm with the
other's direction (here $r_B\,u_A$) and is trained onward; we ask whether its
grokking delay follows the norm donor and its circuit identity follows the
direction donor. The directional dose-response (inset) interpolates the implanted
direction along the geodesic from $u_A$ to $u_B$ at fixed norm, isolating the
angular coordinate.}
\label{fig:schema}
\end{figure}

\paragraph{Circuit-identity metric.}
Following the Fourier account of modular-arithmetic circuits
\citep{nanda2023progress}, we summarize a network's circuit by the normalized
power spectrum of its token-embedding rows. The domain of the transform must
match the task: for addition, circuits are periodic in the token index, so we
take the FFT over rows $0,\dots,p-1$; for multiplication, circuits are periodic
in the \emph{discrete logarithm} of the token (multiplication modulo $p$ is
addition modulo $p-1$ under $t\mapsto\log_g t$ for a primitive root $g$), so we
reorder rows as $[g^0,g^1,\dots,g^{p-2}]$ before the transform. Using the
index-domain transform for multiplication is blind to genuine circuit
differences---two synthetic multiplication circuits at different frequencies
register cosine similarity $0.9998$ under the wrong domain and $0.0000$ under the
correct one. We define \emph{circuit similarity} as the cosine similarity of two
normalized power spectra and, for a chimera, report
$\mathrm{CFS\_lean} = \mathrm{sim}(\text{chimera},B_{\text{final}}) -
\mathrm{sim}(\text{chimera},A_{\text{final}})$, so that negative values indicate
an $A$-like final circuit and positive values a $B$-like one.

\paragraph{Directional dose-response.}
To move continuously from recipient to donor direction we interpolate along the
geodesic on the unit sphere,
\[
\mathrm{slerp}(u_A,u_B,t) = \frac{\sin\big((1{-}t)\phi\big)}{\sin\phi}\,u_A +
\frac{\sin\big(t\phi\big)}{\sin\phi}\,u_B, \qquad \phi = \arccos(u_A\!\cdot u_B),
\]
and implant $r_A\cdot\mathrm{slerp}(u_A,u_B,t)$ at fixed recipient norm $r_A$.
Spherical interpolation preserves unit norm along the entire path, so the
intervention varies \emph{only} the angular coordinate; linear interpolation of
unit vectors would shrink the mixture's norm (up to ${\sim}29\%$ at $t=0.5$ for
near-orthogonal endpoints), simultaneously changing magnitude and direction and
confounding the two axes we seek to dissociate. The endpoint $t{=}1$ coincides
exactly with \textsc{reverse\_radial} and is reused rather than recomputed.

\paragraph{Statistical protocol.}
We select checkpoint pairs so that no seed appears in more than one pair
(disjoint matching), yielding $10$ fully independent pairs per task drawn from
$20$ distinct seeds. This avoids pseudoreplication: reusing a single seed across
many pairs would make pooled statistics non-independent \citep{hurlbert1984}. We
report cluster-level statistics accordingly \citep{cameron2015}, use exact
sign tests for directional consistency, and bootstrap confidence intervals for
pooled means. Decision thresholds separating a usable signal from a confound
were fixed before analysis. Throughout, we distinguish ``statistically
distinguishable from zero'' from ``large''; the delay effect below is the former
but not the latter.

\paragraph{Threshold localization by bisection.}
Under our protocol (full-batch training, reset optimizer) repeated continuations
are sufficiently stable that the interpolation threshold $t^\star$ per pair is
limited by measurement \emph{resolution} rather than by sampling noise. We
therefore localize $t^\star$ by bisection: the coarse grid brackets the sign
change of $\mathrm{CFS\_lean}$ to an interval of width $0.25$, and each bisection
step halves it, reaching $\pm 1/64$ in three evaluations per pair---an order of
magnitude fewer continuations than a uniform fine grid over all pairs. Because
CUDA kernels are not bit-deterministic, we report $t^\star$ with its bisection
half-width as a resolution bound rather than as an exact real number.

\section{The angular component carries transferable circuit identity}
\label{sec:endpoints}

\begin{figure}[t]
\centering
\includegraphics[width=\linewidth]{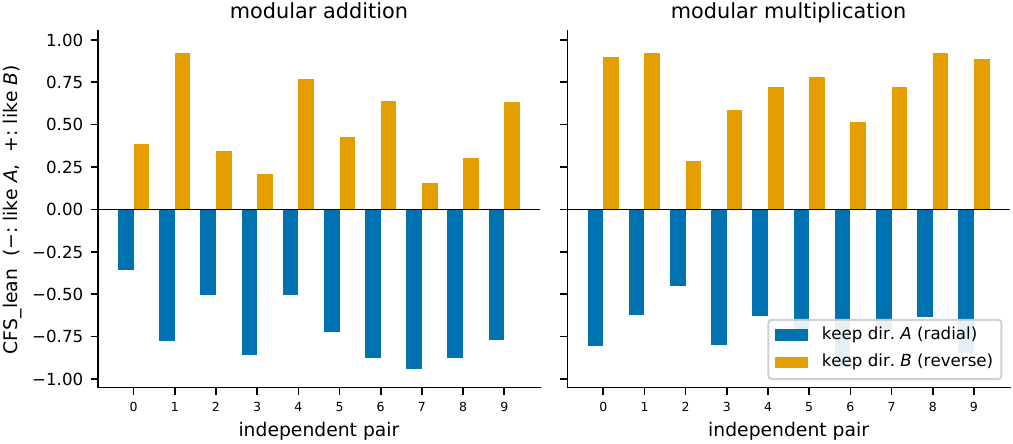}
\caption{\textbf{The angular component carries transferable circuit identity.}
Per-pair $\mathrm{CFS\_lean}$ under the endpoint swaps, for both tasks. The
\textsc{radial} variant (keep recipient direction $u_A$; blue) yields an
$A$-like final circuit in every pair (negative), while \textsc{reverse\_radial}
(keep donor direction $u_B$; orange) yields a $B$-like circuit (positive). Pairs
are fully independent (disjoint seeds). Sign is consistent in $40/40$
recombinations pooled across the two variants and two tasks.}
\label{fig:endpoints}
\end{figure}

\paragraph{Endpoint swaps.}
Implanting the donor's direction while keeping the recipient's norm drives the
continued run toward the donor's eventual circuit. On modular addition, the
\textsc{radial} variant (recipient direction $u_A$, donor norm $r_B$) yields a
final circuit that is $A$-like in every pair (mean $\mathrm{CFS\_lean}=-0.719$,
$10/10$ negative), while its mirror \textsc{reverse\_radial} (donor direction
$u_B$) is $B$-like in every pair (mean $+0.478$, $10/10$ positive). Modular
multiplication replicates this exactly and more strongly on the $B$-side (means
$-0.719$ and $+0.725$; $10/10$ and $10/10$). Pooling the two variants across both
tasks, the final circuit follows the direction donor in $40/40$ independent
recombinations; each per-task, per-variant sign test gives $p=2\!\times\!10^{-3}$
(Figure~\ref{fig:endpoints}).

\paragraph{The effect is donor-specific, not generic perturbation.}
The angle-matched random control---a random direction at the same angular
distance from $u_A$ as $u_B$---produces final circuits with essentially no lean
(mean $-0.031$ on addition, $-0.051$ on multiplication; leans scattered near
zero). Formally, the donor variants are distinguishable from the matched control
on the circuit-identity axis in $58/60$ comparisons on addition and $60/60$ on
multiplication. Thus the transfer reflects the donor's specific direction, not
merely a perturbation of that angular magnitude.

\paragraph{Changing only the norm does not change identity.}
The \textsc{mid\_norm} control (recipient direction, averaged norm) leans
$A$-like exactly as \textsc{radial} does ($-0.707$ vs.\ $-0.719$ on addition;
$-0.736$ vs.\ $-0.719$ on multiplication; $10/10$ negative in both). Altering the
implanted norm while holding direction fixed leaves circuit identity unchanged,
completing the dissociation from the identity side.

\section{Identity transfer is threshold-like, and its threshold tracks recipient norm}
\label{sec:threshold}

\begin{figure}[t]
\centering
\includegraphics[width=\linewidth]{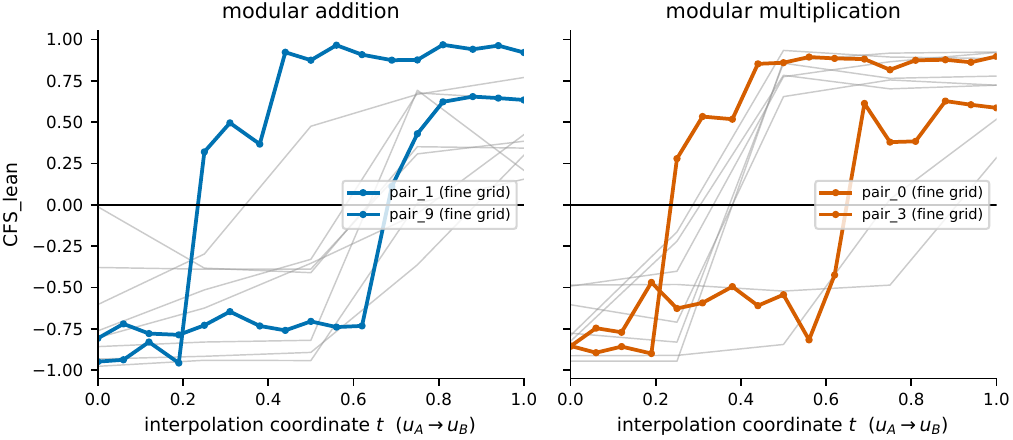}
\caption{\textbf{Identity transfer is threshold-like, not a continuous blend.}
$\mathrm{CFS\_lean}$ as the implanted direction is interpolated along the
geodesic from $u_A$ ($t{=}0$) to $u_B$ ($t{=}1$) at fixed recipient norm. Grey
curves are all pairs on the coarse grid; coloured curves are the representative
pairs measured on a $16$-point fine grid. Each pair stays near one plateau, then
switches over a narrow range---a graded intervention with a sharply graded
response.}
\label{fig:doseresponse}
\end{figure}

\paragraph{A threshold, not a blend.}
Interpolating the implanted direction from $u_A$ to $u_B$ at fixed norm does not
move circuit identity smoothly. Across all $20$ pairs the endpoints have opposite
sign as expected, but the change is concentrated: on average $82$--$88\%$ of the
total change in $\mathrm{CFS\_lean}$ occurs within a single grid step, with
plateaus on either side (Figure~\ref{fig:doseresponse}).

 A $16$-point fine grid
on representative pairs confirms the plateau--jump--plateau shape and agrees with
the bisection threshold to within one grid step. Identity transfer thus behaves
as a switch between two basins rather than a continuous mixture---a graded
intervention with a sharply graded response, consistent with a winner-take-all
selection between distinct circuits.

\begin{figure}[t]
\centering
\includegraphics[width=0.72\linewidth]{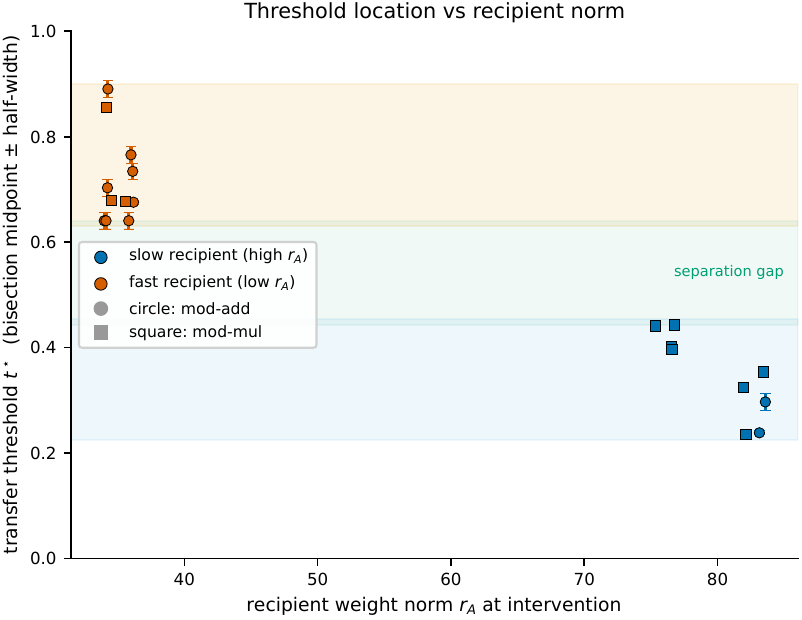}
\caption{\textbf{The transfer threshold is predicted by recipient norm.}
Bisection-localized threshold $t^\star$ (midpoint $\pm$ half-width, $\pm 1/64$)
against the recipient's weight norm $r_A$, for all $20$ pairs across both tasks
(circles: modular addition; squares: modular multiplication). High-norm (slow)
recipients flip early; low-norm (fast) recipients flip late. The two norm
classes separate without overlap; the green band marks the separation gap. The
shaded bands are a visual aid for the grouping, not a fitted relationship.}
\label{fig:tstar}
\end{figure}

\paragraph{The threshold location is predicted by recipient norm.}
The interpolation value $t^\star$ at which identity flips is not constant across
pairs, and it is organized by the recipient's weight norm. Localizing $t^\star$
to $\pm 1/64$ by bisection, pairs whose recipient has high norm (far from
grokking) flip early---only a small move toward $u_B$ is needed---whereas pairs
whose recipient has low norm (near grokking) flip late. On modular addition the
high-norm (slow) recipients have $t^\star\in[0.238,0.297]$ and the low-norm
(fast) recipients $t^\star\in[0.641,0.891]$, a clean gap of $0.34$; on modular
multiplication the corresponding ranges are $[0.235,0.443]$ and $[0.678,0.856]$,
a gap of $0.23$. The two norm classes separate perfectly on every pair in both
tasks; pooling all $20$ pairs preserves perfect separation (gap $0.20$). Treating
each task's separation as an exact permutation test under the observed grouping
gives probabilities $2.2\!\times\!10^{-2}$ and $8.3\!\times\!10^{-3}$, and
$1.9\!\times\!10^{-4}$ jointly (Figure~\ref{fig:tstar}).

 Because the intervention
holds the recipient's norm fixed and varies only direction, this is a statement
about \emph{susceptibility}: at high norm, circuit identity is easily overwritten
by a donor direction; at low norm, it resists.

\paragraph{Scope of the norm--threshold relationship.}
We report this as a separation between two norm classes, not as a fitted
continuous law $t^\star(r)$. The binary slow/fast grouping coincides with our
pair-selection design, which favours large norm differences and therefore
produces two well-separated norm groups rather than a dense sweep; establishing
a functional form would require pairs sampled across a continuum of recipient
norms. What the data support is the ordinal claim---higher recipient norm,
earlier threshold---which holds without exception here.

\section{The radial component: a modest, distributed delay effect}
\label{sec:delay}

The norm carries a genuine but weak influence on grokking \emph{delay}, and no
detectable identity information. Under the \textsc{radial} variant, the chimera's
delay shifts toward the norm donor by a normalized displacement whose per-pair
donor-following score averages $0.28$ on addition ($9/10$ positive) and $0.41$ on
multiplication ($10/10$ positive)---reliably nonzero but far from the value that
would indicate the chimera fully inherits the donor's delay. Expressed as a
fractional displacement between recipient and donor delay, the chimera moves
about $30\%$ of the way toward the donor, pooled across tasks. The three
implanted-norm levels available per pair (recipient, averaged, donor norm) order
delay monotonically in $17/20$ pairs, consistent with a graded but weak
norm--delay relationship. Crucially, when the norm swap is restricted to
individual layer groups the delay effect vanishes for every group---it is a
property of the whole weight vector, not attributable to any single layer.

\begin{figure}[t]
\centering
\includegraphics[width=\linewidth]{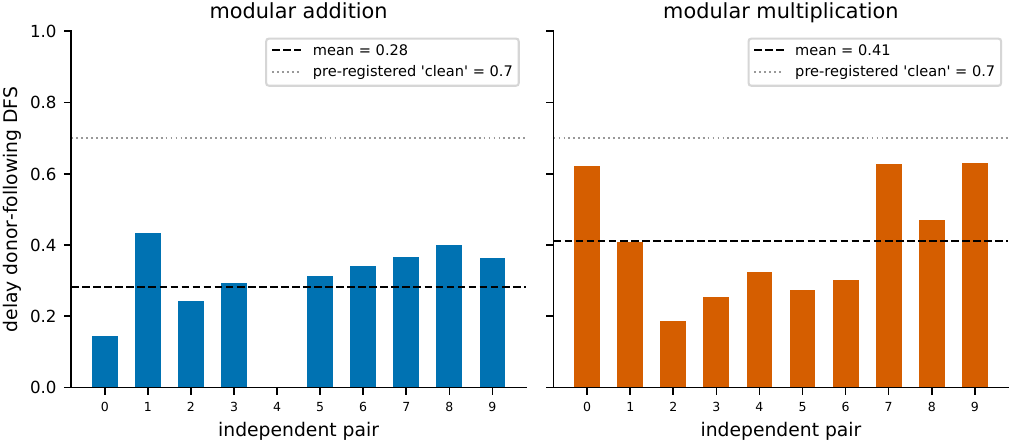}
\caption{\textbf{The radial component carries a modest, distributed delay
effect.} Per-pair delay donor-following score under the \textsc{radial} variant.
The mean is reliably positive (dashed) but no pair reaches the pre-registered
``clean'' threshold of $0.7$ (dotted): the norm shifts delay toward its donor,
but only partially, and the effect vanishes when the swap is restricted to
individual layer groups (not shown).}
\label{fig:delay}
\end{figure}
Relative to the strong, localizable, threshold-structured identity effect, the
delay effect is weak, distributed, and carries no identity signal: the two
geometric components play different roles.

\section{Supporting characterization}
\label{sec:supporting}

\paragraph{Which layers carry the identity signal (summary; full analysis in Appendix~\ref{app:layerwise}).}
Restricting the directional swap to individual layer groups shows that the
identity-carrying signal is not the embedding's alone: on both tasks, the
attention, MLP, and unembedding groups each individually pull the final circuit
toward the donor, but combining them gives a weaker lean than the strongest
single group---the contributions are non-additive. We regard this as
suggestive rather than established, given three pairs per task and a
circularity that affects the embedding-based readout specifically; the full
numbers, the circularity issue, and the figure are in
Appendix~\ref{app:layerwise}.

\paragraph{Circuits keep reorganizing after grokking.}
Tracking each run's spectral similarity to its own final circuit across training,
we find that $50\%$ of the post-grokking reorganization is reached at a median of
$350$ steps after the test-accuracy transition, and $90\%$ at a median of
$1250$--$1850$ steps, in every one of $32$ seeds per task. Circuit structure thus
continues to consolidate well after generalization is behaviourally complete;
our contribution here is to quantify this on independent seeds with a spectral
identity measure, complementing qualitative reports of a post-grokking cleanup
phase \citep{nanda2023progress}.

\paragraph{No two seeds share a circuit.}
At a similarity threshold of $0.9$, all $32$ seeds per task occupy distinct
circuits; mean pairwise similarity is $0.18$--$0.20$, near the $1/\sqrt{d}$
baseline for the fingerprint dimension, and strictly positive for every pair.
Independently trained networks converge to distinct but not unrelated spectral
structures; whether the shared positive component reflects a common substrate is
left open.

\section{Optimizer-state ablation}
\label{sec:optablation}

AdamW carries, in addition to the weights $\theta$, a per-parameter history of
gradient statistics (the first and second moment estimates $m,v$). When we
build a chimera and continue training, this history could in principle belong
to either parent run rather than being a property of $\theta$ itself. We test
this directly: for both the endpoint-swap intervention (Section~\ref{sec:endpoints})
and the threshold-localization intervention (Section~\ref{sec:threshold}), we
repeat the experiment under three sources for the continuation optimizer's
moments---\textsc{reset} (freshly initialized, as reported above),
\textsc{recipient} (transplant $A$'s moments), and \textsc{donor} (transplant
$B$'s moments)---and ask whether the reported effect survives all three.

\paragraph{Circuit-identity transfer (C1) is unaffected.}
Table~\ref{tab:optablation-c1} reports $\mathrm{CFS\_lean}$ for the
\textsc{radial} and \textsc{reverse\_radial} endpoint swaps under all three
optimizer conditions. The sign is consistent in $10/10$ pairs in every cell but
one ($9/10$ for \textsc{radial} on addition under \textsc{recipient}), and mean
magnitudes shift by at most $0.08$ across conditions. The identity-transfer
effect is a property of the weight configuration, not of which run's gradient
history the optimizer happens to carry.

\begin{table}[t]
\centering
\small
\begin{tabular}{llccc}
\toprule
Task & Variant & reset & recipient & donor \\
\midrule
\multirow{2}{*}{mod\_add} & radial          & $-0.719$ (10/10) & $-0.639$ (9/10)  & $-0.719$ (10/10) \\
                          & reverse\_radial & $+0.478$ (10/10) & $+0.539$ (10/10) & $+0.507$ (10/10) \\
\multirow{2}{*}{mod\_mul} & radial          & $-0.719$ (10/10) & $-0.693$ (10/10) & $-0.673$ (10/10) \\
                          & reverse\_radial & $+0.725$ (10/10) & $+0.742$ (10/10) & $+0.723$ (10/10) \\
\bottomrule
\end{tabular}
\caption{Mean $\mathrm{CFS\_lean}$ (sign-correct pairs / total) for the two
endpoint-swap variants under the three optimizer-state conditions. Sign and
magnitude are stable across all conditions.}
\label{tab:optablation-c1}
\end{table}

\begin{figure}[t]
\centering
\includegraphics[width=\linewidth]{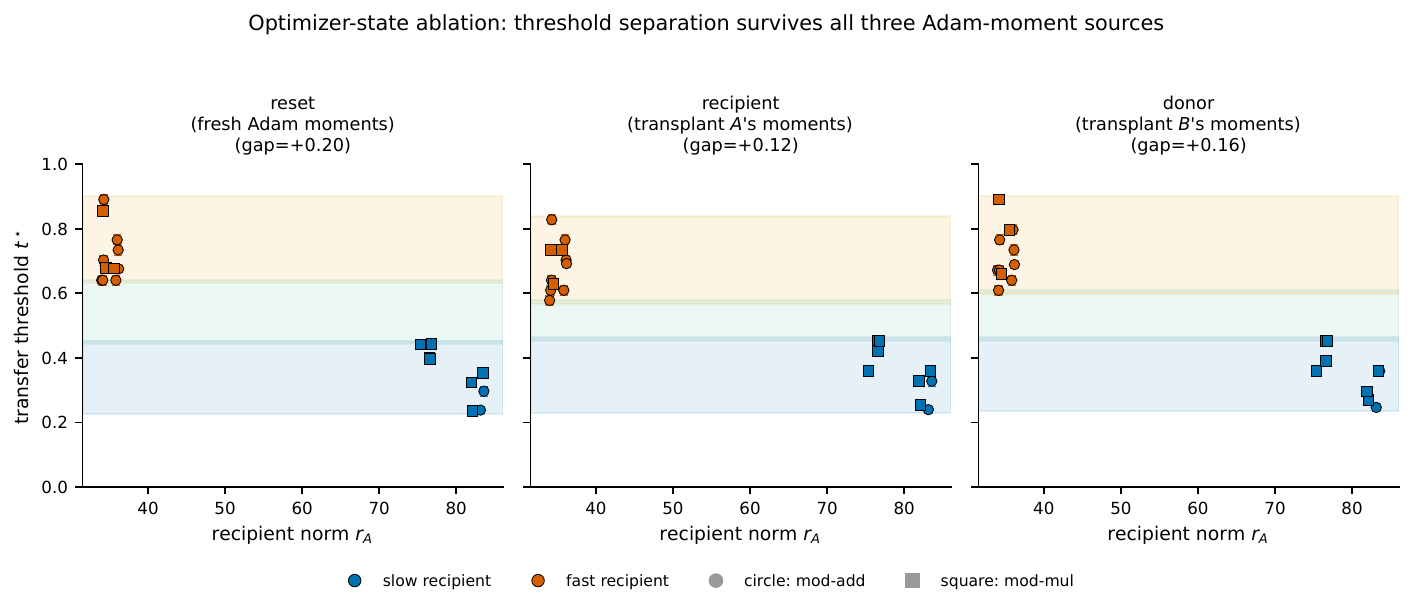}
\caption{\textbf{Optimizer-state ablation for the threshold--norm relationship
(C5).} Bisection-localized $t^\star$ against recipient norm $r_A$, repeated
under three sources for the continuation optimizer's Adam moments: freshly
reset (left, as reported in Figure~\ref{fig:tstar}), transplanted from the
recipient (middle), and transplanted from the donor (right). All $20$ pairs,
both tasks. The slow/fast recipient-norm groups separate without overlap under
every condition; the gap narrows slightly under moment transplantation but
never closes.}
\label{fig:optablation}
\end{figure}

\paragraph{The threshold--norm relationship (C5) is unaffected.}
Repeating the full bisection procedure under \textsc{recipient} and
\textsc{donor} moments, on all $20$ pairs across both tasks, the slow/fast
recipient-norm groups separate perfectly under every one of the three
conditions (Figure~\ref{fig:optablation}). The separation gap narrows somewhat
under moment transplantation (e.g.\ modular addition: $0.34$ under reset versus
$0.25$ under recipient and donor) but never closes, and the per-task exact
permutation probability is unchanged because the same slow/fast grouping
separates in every condition ($2.2\!\times\!10^{-2}$ for addition,
$8.3\!\times\!10^{-3}$ for multiplication, in each of the three conditions).
Treating the six task--condition separations as independent, the joint
probability of this pattern under the null is $6.3\!\times\!10^{-12}$.
Per-pair, $t^\star$ shifts by an average of $0.03$--$0.04$ when moments are
transplanted---about twice the bisection resolution ($\pm 1/64$), and small
relative to the $0.18$--$0.34$ gap that separates the two recipient-norm
groups. We conclude that the threshold's dependence on recipient norm is a
property of $\theta$, with optimizer moments contributing a secondary,
non-decisive perturbation.

\section{Discussion}
\label{sec:discussion}

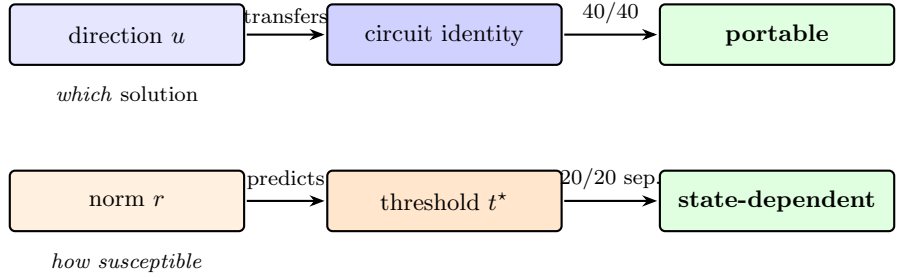
\begin{figure}[t]
\centering
\begin{tikzpicture}[
    >={Stealth[length=2mm]},
    box/.style={rectangle, draw, thick, rounded corners=2pt, minimum width=3.1cm,
                minimum height=8mm, align=center, font=\small},
    lbl/.style={font=\footnotesize, align=center},
    every node/.style={outer sep=1pt},
]
\node[box, fill=blue!10] (dir) at (0,2.2) {direction $u$};
\node[box, fill=blue!18] (id) at (4.2,2.2) {circuit identity};
\node[box, fill=green!12] (port) at (8.6,2.2) {\textbf{portable}};
\draw[->, thick] (dir) -- (id) node[lbl, midway, above] {transfers};
\draw[->, thick] (id) -- (port) node[lbl, midway, above] {$40/40$};

\node[box, fill=orange!12] (norm) at (0,0) {norm $r$};
\node[box, fill=orange!20] (susc) at (4.2,0) {threshold $t^\star$};
\node[box, fill=green!12] (state) at (8.6,0) {\textbf{state-dependent}};
\draw[->, thick] (norm) -- (susc) node[lbl, midway, above] {predicts};
\draw[->, thick] (susc) -- (state) node[lbl, midway, above] {20/20 sep.};

\node[lbl, below=1mm of dir] {\emph{which} solution};
\node[lbl, below=1mm of norm] {\emph{how susceptible}};
\end{tikzpicture}
\caption{\textbf{Division of labour between the two geometric components.}
Direction carries a transferable component associated with circuit identity,
i.e.\ \emph{which} solution a chimera approaches; this is portable across
independent trajectories regardless of recipient state. Norm does not carry
identity, but predicts the interpolation threshold at which a donor's identity
overwrites the recipient's---i.e.\ \emph{how susceptible} the recipient is to
that overwrite, a property of the recipient's dynamical state rather than of
the donor's content.}
\label{fig:conceptual}
\end{figure}

Our results map a division of labour between the two geometric components of a
weight vector (Figure~\ref{fig:conceptual}). Direction indexes \emph{which}
solution a trajectory approaches: transplanting it across independent runs
transfers circuit identity, does so
specifically to the donor's content, and does so as a threshold switch between
basins. Norm plays a different role: it carries a weak, spatially distributed
influence on \emph{when} generalization occurs and no identity information, but
it governs \emph{how susceptible} an identity is to being overwritten---high-norm
recipients flip to a donor's circuit under a small directional nudge, low-norm
recipients resist. Interpreted through the basin picture, a chimera that is far
from convergence (high norm) sits in a shallow region of the landscape where a
modest directional change suffices to cross into a neighbouring basin, whereas a
chimera near convergence (low norm) is already committed. This interpretation is
consistent with the norm--timescale relationship we reported previously for
single-run delay, but we do not claim it explains that relationship.

The non-additivity of the identity signal across hidden layers is an open
puzzle. We considered and ruled out a training-instability explanation---the
combined swaps re-grok to full accuracy rather than destabilizing---leaving a
genuine interaction among layer groups as the more likely account, which we do
not resolve here.

\paragraph{What does not follow from our results.}
Our findings do \emph{not} show that direction alone determines circuit identity;
that norm is irrelevant to representation; that basin geometry is fully
characterized by weight norm; or that the observed dissociation generalizes
beyond the two cyclic-group tasks and single architecture studied here. Each of
these is a plausible next question, not a conclusion we are entitled to.

\section{Limitations}
\label{sec:limitations}

We study a single architecture and two tasks from the same (cyclic-group) family;
whether the dissociation holds for non-group tasks or larger models is untested.
We explored a non-cyclic task (sparse parity) and found that it groks cleanly
only under a different optimization regime (weight decay $0.1$, lower learning
rate); rather than report a cross-task ``replication'' confounded by mismatched
hyperparameters, we leave the non-cyclic extension to future work with the
validated configuration in hand. Our circuit-identity metric is an
embedding-spectrum proxy for the full circuit, not a complete circuit
equivalence, and the embedding-swap localization result is circular for that
reason (Section~\ref{sec:supporting}). The threshold $t^\star$ is localized on a
grid of resolution $\pm1/64$; the norm--threshold relationship is established as a
two-class separation rather than a continuous law, because our pair selection
produces well-separated norm groups by design. Continued-training
hyperparameters are held fixed regardless of donor and recipient origin. The
optimizer-state ablation (Section~\ref{sec:optablation}) covers all $20$ pairs
for the bisection results but only two representative pairs per task for the
underlying fine interpolation grid, matching the scope of the main fine-grid
analysis.

\section{Conclusion}

Cross-trajectory chimera interventions dissociate the roles of weight magnitude
and direction in grokking. Direction carries a transferable, donor-specific
component associated with circuit identity and transfers it as a threshold switch
between basins; the threshold's location is predicted by the recipient's norm,
revealing that transferability depends on the recipient's dynamical state. Norm
carries only a modest, distributed delay effect and no identity signal. The
adaptive bisection procedure we use to localize the threshold is a reusable tool
for interventions that are stable under their training protocol. Beyond the
specific geometry of grokking, the chimera construction offers a general template
for asking not just what a partially trained network contains, but what of it is
portable to another.

\subsubsection*{Reproducibility Statement}
All experiments use publicly reproducible task definitions, fixed random seeds
recorded per run, and a pipeline that regenerates every table and figure from raw
logs. The full analysis (pair selection, chimera and interpolation
interventions, bisection, and statistical reporting) is released as scripts with
per-job resume markers; each reported number is emitted directly by the analysis
code from the experiment CSVs.

\bibliographystyle{plainnat}
\bibliography{references}

\appendix
\section{Layer-localization of the identity signal}
\label{app:layerwise}

\begin{figure}[h]
\centering
\includegraphics[width=\linewidth]{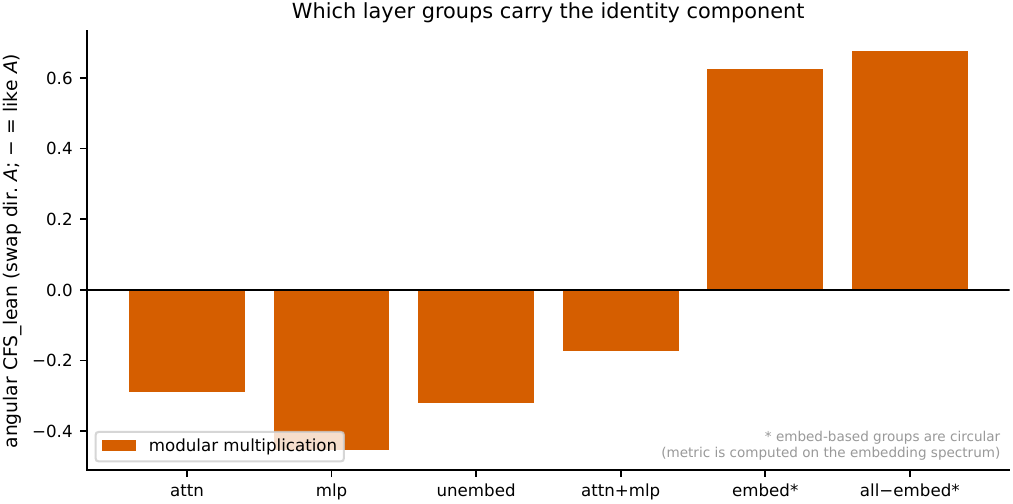}
\caption{\textbf{The identity component is carried by multiple hidden groups,
non-additively.} Angular $\mathrm{CFS\_lean}$ when the directional swap is
restricted to one layer group (mean over three pairs). Attention, MLP, and
unembedding each individually carry the correct-sign signal, but combining them
(``attn+mlp'') gives a weaker lean than the strongest single group---the
contributions are non-additive. The embedding-based groups (marked $*$) are
circular because the circuit metric is computed on the embedding spectrum, and
are not interpreted.}
\label{fig:layerwise}
\end{figure}

This appendix gives the full layer-group analysis summarized in
Section~\ref{sec:supporting}. Restricting the \emph{directional} swap to
individual layer groups shows that the identity-carrying signal is not the
embedding's alone. On modular addition, the attention, MLP, and unembedding
groups each individually pull the final circuit toward the donor
($\mathrm{CFS\_lean}\approx-0.6$ each when swapped in the $A$-direction test),
yet swapping all of them together produces a markedly weaker lean ($-0.145$):
the contributions are non-additive. Modular multiplication reproduces the
qualitative pattern---each hidden group carries the correct-sign signal
(attention $-0.29$, MLP $-0.45$, unembedding $-0.32$) and the combination is
again non-additive---though per-task magnitudes differ (Figure~\ref{fig:layerwise}).

We flag one confound explicitly: because the circuit metric is computed
\emph{on} the embedding spectrum, directly swapping the embedding's direction
manipulates the very object being measured, so the embedding and
``all-but-embedding'' results are circular and we do not interpret them; the
non-circular hidden-layer results are the basis for the redundancy claim above,
which we regard as suggestive rather than established, given three pairs per
task.

\end{document}